\documentclass{article}
\usepackage[T1]{fontenc}
\usepackage[utf8]{inputenc}
\usepackage{float}
\usepackage{amsmath}
\usepackage{graphicx}
\usepackage{microtype}
\usepackage[unicode=true,
 bookmarks=false,
 breaklinks=false,pdfborder={0 0 1},backref=section,colorlinks=false]
 {hyperref}

\makeatletter

\floatstyle{ruled}
\newfloat{algorithm}{tbp}{loa}
\providecommand{\algorithmname}{Algorithm}
\floatname{algorithm}{\protect\algorithmname}

\usepackage{arxiv}

\usepackage{url}
\usepackage{booktabs}
\usepackage{amsfonts}
\usepackage{nicefrac}
\usepackage[noend]{algpseudocode}
\usepackage{algorithmicx}
\usepackage{algorithm}\usepackage{threeparttable}
\usepackage{cite}
\usepackage{url}

\title{A short note on the decision tree based \\ neural turing machine}
\author{%
Yingshi Chen \\  
  \texttt{gsp@grusoft.com} \\
}

\@ifundefined{showcaptionsetup}{}{%
 \PassOptionsToPackage{caption=false}{subfig}}
\usepackage{subfig}
\makeatother

\begin{document}
\title{A short note on the decision tree based neural turing machine}
\maketitle
\begin{abstract}
Turing machine and decision tree have developed independently for
a long time. With the recent development of differentiable models,
there is an intersection between them. Neural turing machine(NTM)
opens door for the memory network. It use differentiable attention
mechanism to read/write external memory bank. Differentiable forest
brings differentiable properties to classical decision tree. In this
short note, we show the deep connection between these two models.
That is: differentiable forest is a special case of NTM. Differentiable
forest is actually decision tree based neural turing machine. Based
on this deep connection, we propose a response augmented differential
forest (RaDF). The controller of RaDF is differentiable forest, the
external memory of RaDF are response vectors which would be read/write
by leaf nodes. 
\end{abstract}

\section{Introduction}

Neural Turing Machine(NTM) \cite{graves2014neural,paassen2020reservoir,collier2018implementing,graves2016hybrid}
enhance classical turing machine with differentiable attention mechanism
to visit datas/programs in the external memory bank. NTM has different
names in different papers, for example, memory augmented neural networks(MANN),
reservoir memory machines, differentiable neural computer... These
names reveal the interesting aspects of NTM model in different ways.
With external “artificial” working memory, NTM would have strong capacity
to store and retrieve pieces of information just like human being.
Some experiments in \cite{graves2014neural} verified its defectiveness
over famous LSTMs models. Since the pioneering work of \cite{graves2014neural},
NTM has been deeply studied and applied to many applications, including
machine translation, recommendation systems, slam, reinforcement learning,
object tracking, video understanding, Graph Networks, etc \cite{santoro2016meta,parisotto2017neural,chen2018sequential,rae2016scaling,zhang2017neural,ebesu2018collaborative,yang2018learning,na2017read,gulcehre2018dynamic,pham2018graph,kenter2017attentive}.
All the controller in these works are neural networks. Our work is
the first proposition of differentiable decision tree based NTM.

Differentiable decision tree\cite{kontschieder2015deep,popov2019neural}
brings differentiable properties to classical decision tree. Compared
with widely used deep neural networks, tree model still has some advantages.
Its structure is very simple, easy to use and explain the decision
process. Especially for tabular data, gradient boosted decision trees(GBDT)\cite{friedman2001greedy}
models usually have better accuracy than deep networks, which is verified
by many Kaggle competitions and real applications. But the classical
tree-based models lack of differentiability, which is the key disadvantage
compare to the deep neural networks. Now the differentiable trees
also have full differentiability. So we could train it with many powerful
gradient-based optimization algorithms (SGD, Adam,…). We could use
batch training to reduce memory usage greatly. And we could use the
end-to-end learning to reduce many preprocess works. The differential
operation would search in a larger continuous space, so it can find
more accurate solutions. The experiments on some large datasets showed
differentiable forest model has higher accuracy than best GBDT libs
(LightGBM, Catboost, and XGBoost)\cite{chen2020attention}.

In recent years, different research teams\cite{kontschieder2015deep,popov2019neural,tanno2018adaptive,yang2018deep,lay2018random,feng2018multi,silva2019optimization,hazimeh2020tree}
have proposed different models and algorithms to implement the differentiability.\cite{kontschieder2015deep}
is a pioneering work in the differentiable decision tree model. They
present a stochastic routing algorithm to learn the split parameters
via back propagation. The tree-enhanced network gets higher accuracy
than GoogleNet \cite{szegedy2015going}, which was the best model
at the time of its publication. \cite{popov2019neural} introduces
neural oblivious decision ensembles (NODE) in the framework of deep
learning. The core unit of NODE is oblivious decision tree, which
uses the same feature and decision threshold in all internal nodes
of the same depth. There is no such limitation in our algorithm. As
described in section \ref{sec:algorithm}, each internal node in our
model could have independent feature and decision threshold. \cite{tanno2018adaptive}
presents adaptive neural trees (ANTs) constructed on three differentiable
modules(routers, transformers and solvers). In the training process,
ANTs would adaptively grow the tree structure. Their experiment showed
its competitive performance on some testing datasets. \cite{yang2018deep}
presents a network-based tree model(DNDT). Their core module is soft
binning function, which is a differentiable approximation of classical
hard binning function. \cite{lay2018random} propose random hinge
forests or random ferns with differentiable ReLU-like indicator function.
Then the loss function would be optimized end-to-end with stochastic
gradient descent algorithm. 

It's clear that these groups did not realize the connection with NTM.
Our work in section \ref{sec:radf} reveals that differentiable forest
is a special case of NTM. This would help to improve the differentiable
forest model.

\section{Differentiable forest based neural turing machine \label{sec:radf} }

In this section, we would first give the main structure/algorithm
of differentiable forest and NTM, then reveal the essential connection
between them. That is, differentiable forest is also neural turing
machin with specific controller and attention mechanism. To our knowledge,
this is the first proposition of differentiable forest based NTM.

\subsection{Neural Turing Machine}

\label{subsec:neural-turing-machine} 

Just like the classic turing machine, NTM is a recurrent machine with
two main modules, the first is the controller and the second is external
memory bank $\mathbf{M}$. The external memory bank $\mathbf{M}$
is usually defined as $N\times W$ matrix, which contains $N$cells
(memory locations) and the size of each cell $\mathbf{M^{\mathit{i}}}$
is $W$. The controller would read and write $\mathbf{M}$ to update
its state. The main innovation of NTM is its attention mechanism which
would update the weights of each read/write operation. 

At each time step $t$, NTM would update the weight $w_{t}^{i}$ of
each cell. Then the read operation would get $\mathbf{r}^{t}$ and
write operation would update each cell $\mathbf{M^{\mathit{i}}}$
as :

\begin{equation}
\begin{split}\mathbf{r_{\mathrm{\mathit{t}}}} & =\sum_{i=1}^{N}w_{t}^{i}\mathbf{M}_{t}^{i}\\
\mathbf{M}_{t}^{i} & =\mathbf{M}_{t-1}^{i}\left[1-w_{t}^{i}\mathbf{e}_{t}\right]+w_{t}^{i}\mathbf{a}_{t}
\end{split}
\label{F:iteration of NTM}
\end{equation}

where $\mathbf{e}_{t}$ is additional erase vector and $\mathbf{a}_{t}$
is add vector. For more detail of practical robust implementation
of NTM, please see \cite{collier2018implementing}.

\subsection{Response augmented differential forest (RaDF)}

The response augmented differential forest $\mathbf{RaDF(T,Q)}$ has
two main modules. The first is controller $\mathbf{T}$ which has
$K$ differentiable decision trees $\left[T_{1},T_{2},\cdots,T_{K}\right]$;
the second is external memory bank $\mathbf{Q}$. Each cell of $\mathbf{Q}$
is actually response corresponding to some leaf nodes\cite{chen2020attention}.
The controller $\mathbf{T}$ would read an write response bank $\mathbf{Q}$.
Each leaf node is just the head in the NTM to read/write response
bank $\mathbf{Q}$. So the response augmented differential forest
$\mathbf{DF(T,R)}$ is just a specific NTM.

\begin{figure}[H]
\centering \subfloat[Simplest response augmented differential forest with only three nodes
(one root node with two child nodes)]{{\includegraphics[width=5cm]{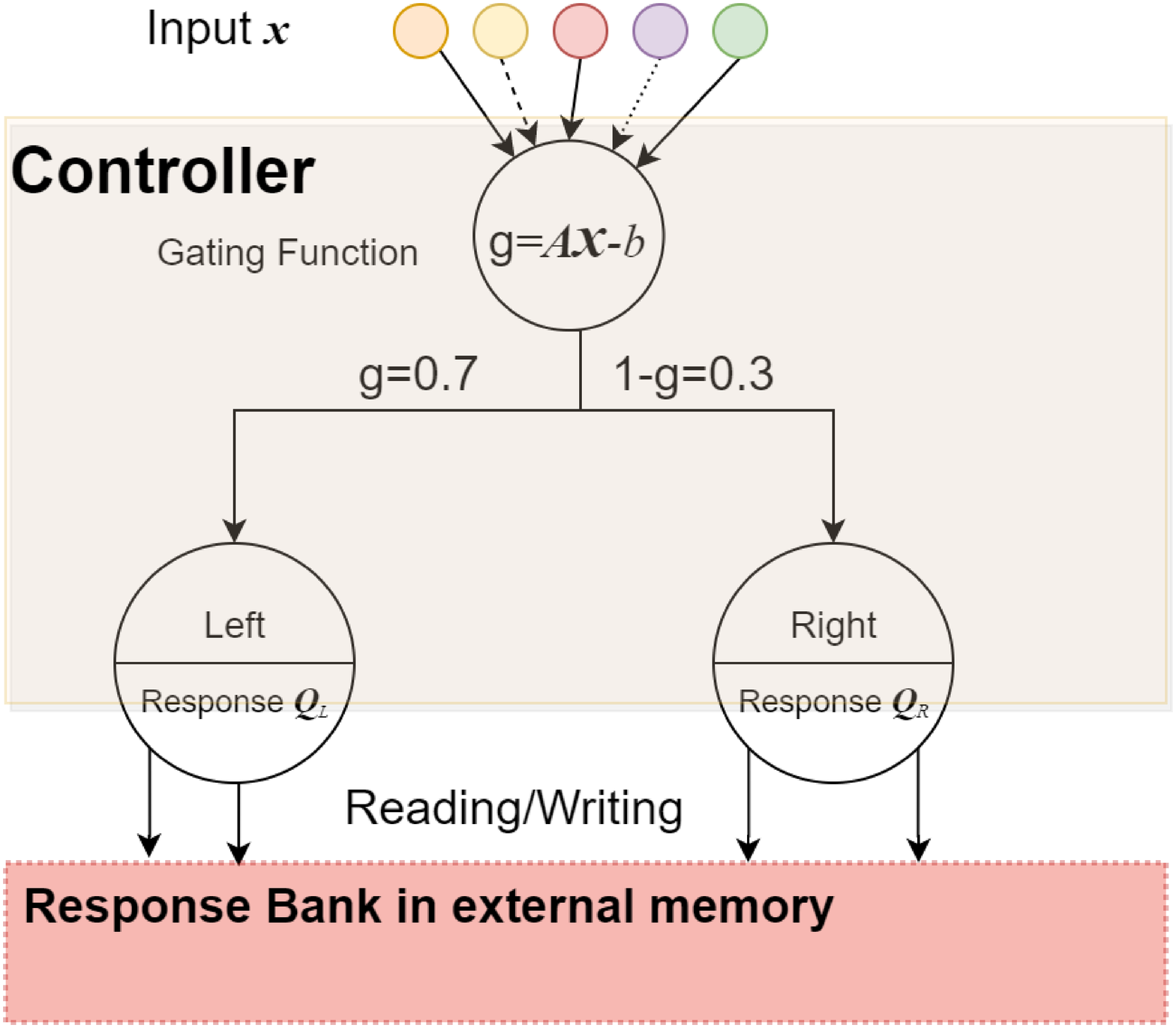} }

}\qquad{}\subfloat[Response augmented differential forest and its response bank. In this
sample, The input $x$ would reach $n_{3}$ with probability $1-g_{1}$
and reach $n_{6}$ with probability $(1-g_{1})g_{3}$]{{\includegraphics[width=5cm]{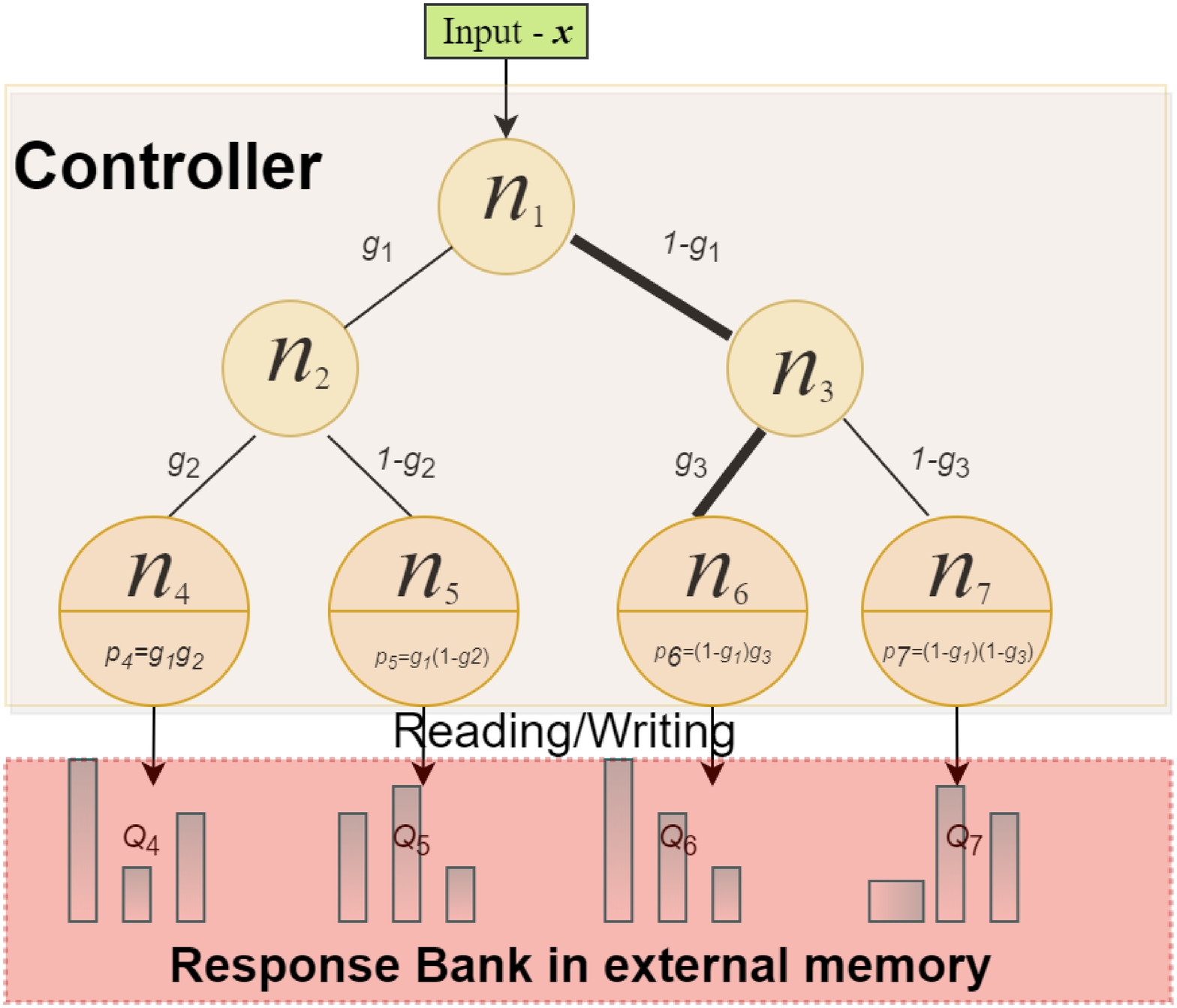} }

}\caption{Differentiable tree with response in outer memory}
\label{fig:trees} 
\end{figure}

For a dataset with N samples $\boldsymbol{X}=\left\{ x\right\} $
and its target $\boldsymbol{Y}=\left\{ y\right\} $. Each $x$ has
M attributes, $x=\left[x_{1},x_{2},\cdots,x_{M}\right]^{T}$. The
$\mathbf{RaDF(T,Q)}$ would learn $K$ differentiable decision trees
$\left[T_{1},T_{2},\cdots,T_{K}\right]$ and the response bank $\mathbf{Q}$
to minimize the loss between the target $y$ and prediction $\hat{y}$.
\begin{equation}
\hat{y}=\frac{1}{K}\sum_{h=1}^{K}T^{h}\left(x\right)
\end{equation}
Figure \ref{fig:trees}.(a) shows the simplest case of RaDF. The controller
is just a simple decision tree(one root node,two leaf nodes). All
leaf nodes in the controller would read/write corresponding response
in the external memory. For each input $x$, the gating function $g$
at root node would generate probabilities for both leaf nodes. Formula
\ref{F:gate} gives the general definition of gating function $g$
with learn-able parameters $A$ and threshold $b$. $\sigma$ would
map $Ax-b$ to probability between {[}0,1{]}, for example, the sigmoid
function. 
\begin{equation}
g\left(A,x,b\right)=\sigma\left(Ax-b\right)\label{F:gate}
\end{equation}
So as shown in Figure \ref{fig:trees}.(b), The sample $x$ would
be directed to each nodal $j$ with probability $p_{j}$. And finally,
the input $x$ would reach all leaves. For a tree with depth $d$,
we represent the path as ${n_{1},n_{2},\cdots,n_{d}}$, where $n_{1}$
is the root node and $n_{d}$ is the leaf node $j$. $p_{j}$ is just
the product of the probabilities of all nodes in this path: 
\begin{equation}
p_{j}=\prod_{n\in\left\{ n_{1},\cdots,n_{d}\right\} }^ {}g_{n}\label{F:p}
\end{equation}
In the model of classical decision tree, the gating function $g$
is just the heave-side function, either left or right. So for each
sample $x$, only one leaf node is activated in the classical decision
tree, while in RaDF model, all leaf nodes are activated(with probability
$p_{j}$) to read/write response bank $\mathbf{Q}$.

The read operation of leaf node $j$ would return a response vector
$\boldsymbol{q}_{j}=\left(q_{j_{1}},q_{j_{2}},\cdots,q_{j_{F}}\right)^{T}$
. Then the output of tree $h$ is just the probability average of
these responses. 
\begin{equation}
Q_{h}\left(x\right)=\sum_{j\ is\ leaf\ of\ h}p_{j}\boldsymbol{q}_{j}\label{F:leaf_y}
\end{equation}

A single tree is a very weak learner, so we should merge many trees
to get higher accuracy, just like the random forest or other ensemble
learning method. The final prediction $y$ is weighted summary of
all trees. In the simplest case, the weight is always $1/K$, $\hat{y}$
is just the average result. 
\begin{equation}
\hat{y}\left(x\right)=\frac{1}{K}\sum_{h=1}^{K}Q_{h}\left(x\right)\label{F:Y}
\end{equation}

Let $\Theta$ represents all parameters $(A,b,Q)$, then the final
loss would be represented by the general form of formula \ref{F:loss}

\begin{equation}
L(\Theta:x,y)=\frac{1}{K}\sum_{h=1}^{K}L_{h}(\Theta:x,y)=\frac{1}{K}\sum_{h=1}^{K}L_{h}(A,b,Q:x,y)\label{F:loss}
\end{equation}

where $L:\mathbb{R}^{F}\mapsto\mathbb{R}$ is a function that maps
vector to object value. In the case of classification problem, the
classical function of $L$ is cross-entropy. For regression problem,
$L$ maybe mse, mae, huber loss or others. To minimize the loss in
formula \ref{F:loss}, we use stochastic gradient descent(SGD) method
to train this model.

\subsection{Algorithm of response augmented differentiable forest}

As a specified case of nTM, RaDF would also be trained by SGD just
like neural network. The main difference in the implementation is
the update process of controller.

Based on the general loss function defined in formula \ref{F:learn_batch},
we use stochastic gradient descent method \cite{kingma2014adam,ma2018quasi}
to reduce the loss. As formula \ref{F:learn_batch} shows, update
all parameters $\Theta$ batch by batch: 
\begin{equation}
\Theta^{t+1}=\Theta^{t}-\eta\frac{\partial L}{\partial\Theta}\sum_{\left(x,y\right)\in\mathfrak{B}}\frac{\partial L}{\partial\Theta}\left(\Theta^{t};x,y\right)\label{F:learn_batch}
\end{equation}
where $\mathfrak{B}$ is the current batch, $\eta$ is the learning
rate, $(x,y)$ is the sample in current batch.

The detailed algorithm is as follows:

\begin{algorithm}[H]
\caption{Implementation of response augmented differentiable forest}
\hspace*{0.02in} \textbf{Input:} input training, validation and test
dataset \\
 \hspace*{0.02in} \textbf{Output:} learned model: response bank $\mathbf{Q}$
in external memory and threshold values $b$ 

\begin{algorithmic}[1]

\State Init feature weight mattrix $A$ 

\State Init response bank $\mathbf{Q}$ at external memory. Each
leaf node has its response stored in one cell of $\mathbf{Q}$.

\State Init threshhold values $b$ 

\State Init time step $t=0$ 

\While{not converge} 

\For{each batch} \State Calculate gating value at each internal
node

\State \quad{}$g\left(A,x,b\right)=\sigma\left(Ax-b\right)$ 

\State Calculate probability at each leaf node

\State \quad{}$p_{j}=\prod_{n\in\left\{ n_{1},\cdots,n_{d}\right\} }^ {}g_{n}$

\State Read response bank $\mathbf{Q}$, update prediction of each
tree 

\State \quad{}$\hat{y}_{h}\left(x\right)=\sum_{j\ is\ leaf}p_{j}Q_{j}$

\State Calculate the loss $L$

\State Backpropagate to get the gradient

\State \quad{}$\left(\delta A,\delta b,\delta Q\right)\Leftarrow\delta L$ 

\State Update weight of each response cell $\mathbf{\mathit{w_{t}^{i}}}$

\State Update response bank $\mathbf{Q}$ with erase vector $\mathbf{e}_{t}$and
add vector $\mathbf{a}_{t}$

\State $\mathbf{\quad Q}_{t}^{i}=\mathbf{Q}_{t-1}^{i}\left[1-w_{t}^{i}\mathbf{e}_{t}\right]+w_{t}^{i}\mathbf{a}_{t}$

\State Update the parameters: $A$, $b$

\EndFor 

\State Evalue loss at validation dataset

\State $t=t+1$ 

\EndWhile 

\State \Return learned model 

\end{algorithmic} 
\end{algorithm}

\section{Conclusion}

In this short note, we revealed the deep connection between differentiable
forest and neural turing machine. Based on the detailed analysis of
both models, the Response augmented differential forest (RaDF) is
actually a special case of NTM. The controller of RaDF is differentiable
forest, the external memory cells of RaDF are response vectors which
would be read/write by leaf nodes. This novel discovery will deepen
the understanding of both two models and inspire some new algorithms.
We give a detailed training algorithm of RaDF. We will give more detailed
experiments in later papers.

\end{document}